\documentclass[runningheads]{llncs}
\usepackage{graphicx}

\usepackage{todonotes}
\usepackage{hyperref}
\usepackage{multirow}
\usepackage{breakcites}
\usepackage{marvosym}

\newcommand{\concept}[1]{\textsc{#1}}

\begin{document}
\title{Grounding Psychological Shape Space in Convolutional Neural Networks}
%
%
\author{Lucas Bechberger\inst{1}\orcidID{0000-0002-1962-1777} \and
Kai-Uwe K\"uhnberger\inst{1}\orcidID{0000-0003-1626-0598}}
\authorrunning{L. Bechberger and K.-U. K\"uhnberger}
%
\institute{Insititute of Cognitive Science, Osnabr\"uck University \email{\{lbechberger,kkuehnbe\}@uos.de}}
\maketitle              
\begin{abstract}
Shape information is crucial for human perception and cognition, and should therefore also play a role in cognitive AI systems. We employ the interdisciplinary framework of conceptual spaces, which proposes a geometric representation of conceptual knowledge through low-dimensional interpretable similarity spaces. These similarity spaces are often based on psychological dissimilarity ratings for a small set of stimuli, which are then transformed into a spatial representation by a technique called multidimensional scaling. Unfortunately, this approach is incapable of generalizing to novel stimuli. In this paper, we use convolutional neural networks to learn a generalizable mapping between perceptual inputs (pixels of grayscale line drawings) and a recently proposed psychological similarity space for the shape domain. We investigate different network architectures (classification network vs. autoencoder) and different training regimes (transfer learning vs. multi-task learning). Our results indicate that a classification-based multi-task learning scenario yields the best results, but that its performance is relatively sensitive to the dimensionality of the similarity space.

\end{abstract}

\section{Introduction}
\label{Intro}

Shape information plays an important role in human perception and cognition, and can be viewed as a bootstrapping device for constructing concepts \cite{Diesendruck2003,Jones1993,Landau1998}. Based on the principle of cognitive AI \cite{Lieto2021,Marcus2019}, also artificial agents should be equipped with a human-like representation of shapes.

In this paper, we employ the cognitive framework of conceptual spaces \cite{Gardenfors2000}, which proposes a geometric representation of conceptual knowledge based on psychological similarity spaces. It offers a way of neurosymbolic integration \cite{Garcez2015,Maruyama2021} by using an intermediate level of representation between the connectionist and the symbolic approach. The overall conceptual space is structured into different cognitive domains (such as \concept{color} and \concept{shape}), which are represented by low-dimensional psychological similarity spaces with cognitively meaningful dimensions. Conceptual spaces have seen a wide variety of applications in artificial intelligence, linguistics, psychology, and philosophy \cite{Kaipainen2019,Zenker2015}.
Typically, the structure of a conceptual space is obtained based on dissimilarity ratings from psychological experiments, which are then translated into a spatial representation through multidimensional scaling \cite{Borg2005}. In this paper, we consider a recently proposed similarity space for the cognitive domain of shapes \cite{Bechberger2020ICCS,Bechberger2021Shapes}.

The similarity spaces obtained by multidimensional scaling are not able to generalize to unseen inputs -- a novel stimulus can only be mapped into the similarity space after eliciting further dissimilarity ratings \cite{Battleday2021}. In order to generalize beyond the initial stimulus set (which is necessary in practical AI applications), we have recently proposed a hybrid approach \cite{Bechberger2021NOUN}: Psychological dissimilarity ratings are used to initialize the similarity space, and a mapping from image stimuli to coordinates in this similarity space is then learned with convolutional neural networks. Both our own prior study \cite{Bechberger2021NOUN} and related studies by Sanders and Nosofsky \cite{Sanders2018,Sanders2020} used a classification-based transfer learning approach on relatively unstructured similarity spaces involving multiple cognitive domains. In contrast to that, the present study focuses on the single cognitive domain of shapes and investigates a larger variety of machine learning setups, comparing two network types (classification network vs. autoencoder) and two learning regimes (transfer learning vs. multi-task learning).\\

The remainder of this article is structured as follows: In Section \ref{Background}, we provide some general background on convolutional neural networks, conceptual spaces, and the cognitive domain of shapes. We then describe our general experimental setup in Section \ref{Methods}, before presenting the results of our machine learning experiments in Section \ref{Exp}. Finally, Section \ref{Conclusion} summarizes the main contributions of this article and provides an outlook towards future work. All of our results as well as source code for reproducing them are publicly available on GitHub \cite{Bechberger2021GitHub}.\footnotemark \footnotetext{See \url{https://github.com/lbechberger/LearningPsychologicalSpaces/}.}

\section{Background}
\label{Background}

Our work combines the cognitive framework of conceptual spaces \cite{Gardenfors2000} with modern machine learning techniques in the form of convolutional neural networks. In Section \ref{Background:CNN}, we briefly introduce convolutional neural networks, before giving an overview of the conceptual spaces framework in Section \ref{Background:CS}. Section \ref{Background:Shapes} then discusses the cognitive domain of shapes.

\subsection{Convolutional Neural Networks}
\label{Background:CNN}

Artificial neural networks (ANNs) consist of a large number of interconnected units. Each unit computes a weighted sum of its inputs, which is then transformed with a nonlinear activation function. The trainable parameters of an ANN correspond to the weights of these connections and are typically optimized through gradient descent, minimizing the prediction error of the network on a given data set. 

\begin{figure}[t]
	\centering
	\includegraphics[width = \columnwidth]{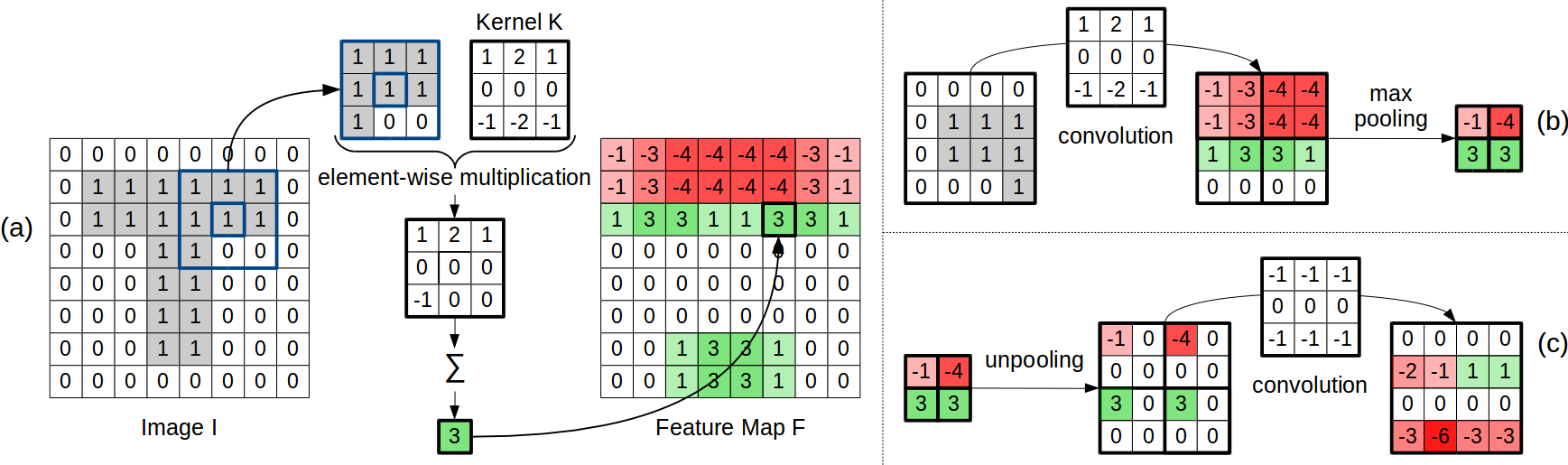}
	\caption{(a) Two-dimensional convolution with a $3\times3$ kernel. (b) Combination of convolution and max pooling. (c) Combination of unpooling and convolution.}
	\label{fig:convolution}
\end{figure}

With respect to computer vision tasks such as image classification, convolutional neural networks (CNNs) are considered to be the most successful ANN variant \cite[Chapter 9]{Goodfellow2016}. They make use of so called \emph{convolutional layers} which apply the same set of weights (represented as kernel $K$) at all locations (see Figure \ref{fig:convolution}a). This and the relatively small size of the kernel (and thus the receptive field of each unit) drastically reduces the number of connections between subsequent layers. CNNs furthermore use so called \emph{max pooling} layers (see Figure \ref{fig:convolution}b) to reduce the size of the image by replacing the output at a certain location by the maximum of its local neighborhood. For a max pooling layer, one has to specify both the pool width (i.e., the size of the area to aggregate over) and the so called ``stride'' (i.e., the step size between two neighboring centers of pooling). 

Typical convolutional networks start from a very high-dimensional input (namely, images) and reduce the representation size in multiple steps until a fairly small representation is reached which can then be used for classification through a softmax layer. However, in some settings one is also interested in the opposite direction: Creating a high-dimensional image from a low-dimensional hidden representation. For instance, autoencoders \cite[Chapter 14]{Goodfellow2016} are an important unsupervised neural network architecture and are commonly used for dimensionality reduction and feature extraction. Autoencoders are typically trained on the task of reconstructing their input at the output layer, using only a relatively low-dimensional internal representation. They consist of an encoder (which compresses information) and a decoder (which reconstructs the original input).

For the encoder, a regular CNN can be used, whose max pooling layers however create a loss of information \cite[Section 20.10.6]{Goodfellow2016}: In Figure \ref{fig:convolution}b, we only keep the maximum value for each $2\times2$ patch of the feature map. Since three out of the four values are discarded completely, it is impossible to accurately reconstruct them. In the decoder, one therefore needs to approximate the inverted pooling function with so called \emph{unpooling} steps. In most cases, one simply replaces each entry of the feature map by a block of size $s \times s$, where the original value is copied to the top left corner and all other entries of the block are set to zero \cite{Dosovitskiy2015} (cf. Figure \ref{fig:convolution}c). Using such an unpooling step followed by a convolution (which is together often called an \emph{upconvolutional} layer) can be seen as an approximate inverse of computing a convolution and a subsequent pooling \cite{Dosovitskiy2015}. This allows us to increase the representation size inside the decoder in order to reconstruct the original input image.

\subsection{Conceptual Spaces}
\label{Background:CS}

A conceptual space as proposed by G\"ardenfors \cite{Gardenfors2000} is a similarity space spanned by a small number of interpretable, cognitively relevant quality dimensions (e.g., \concept{temperature}, \concept{time}, \concept{hue}, \concept{pitch}). One can measure the difference between two observations with respect to each of these dimensions and aggregate them into a global notion of semantic distance. Semantic similarity is then defined as an exponentially decaying function of distance.

The overall conceptual space can be structured into so called domains, which represent, for example, different perceptual modalities such as \concept{color}, \concept{shape}, \concept{taste}, and \concept{sound}. The \concept{color} domain, for instance, can be represented by the three dimensions \concept{hue}, \concept{saturation}, and \concept{lightness}, while the \concept{sound} domain is spanned by the dimensions \concept{pitch} and \concept{loudness}. Based on psychological evidence \cite{Attneave1950,Shepard1964}, distance within a domain is measured with the Euclidean metric, while the Manhattan metric is used to aggregate distances across domains.

G\"ardenfors defines properties like \concept{red}, \concept{round}, and \concept{sweet} as convex regions within a single domain (namely, \concept{color}, \concept{shape}, and \concept{taste}, respectively). Concept hierarchies are an emergent property of this spatial representation, such as the \concept{sky blue} region being a subset of the \concept{blue} region.
Based on properties, G\"ardenfors now defines full-fleshed concepts like \concept{apple} or \concept{dog} by using one convex region per domain, a set of salience weights (which represent the relevance of the given domain to the given concept), and information about cross-domain correlations. The \concept{apple} concept may thus be represented by regions for \concept{red}, \concept{round}, and \concept{sweet} in the domains of \concept{color}, \concept{shape}, and \concept{taste}, respectively. This geometric representation of knowledge enables a straightforward implementation of common-sense reasoning strategies such as interpolative and extrapolative reasoning \cite{Derrac2015,Schockaert2011}. Moreover, it can be liked to the prototype theory of concepts from psychology \cite{Rosch1976} by relating typicality to the distance from the center of a conceptual region.

A popular way of obtaining a conceptual similarity space is based on psychological dissimilarity ratings \cite{Gardenfors2000}. These dissimilarity ratings are collected for a fixed set of stimuli in a psychological experiment. They are then converted into an $n$-dimensional geometric representation of the stimulus set by using a technique called ``multidimensional scaling'' (MDS), which ensures that geometric distances between pairs of stimuli reflect their psychological dissimilarity \cite{Borg2005}. The similarity spaces produced by MDS do not readily generalize to unseen stimuli: Mapping a novel input into the similarity space requires one to collect additional similarity ratings and then to re-run the MDS algorithm on the enlarged dissimilarity matrix \cite{Battleday2021}.

Artificial neural networks (ANNs) on the other hand are capable of generalizing beyond the training examples, but often use relatively high-dimensional internal representations which are hard to interpret and not necessarily psychologically grounded. In our proposed hybrid approach \cite{Bechberger2021NOUN}, we therefore use MDS on human dissimilarity ratings to ``initialize'' the similarity space and ANNs to learn a mapping from stimuli into this similarity space, where the stimulus-point mappings are treated as labeled training instances for a regression task. In general, ANNs require large amounts of data to optimize their weights, but the number of stimuli in a psychological study is necessarily small. We propose to resolve this dilemma not only through data augmentation (i.e., by creating additional inputs through minor distortions), but also by introducing an additional training objective (e.g., correctly classifying the given images into their respective classes). This additional training objective can also be optimized on additional stimuli that have not been used in the psychological experiment. Using a secondary task with additional training data constrains the network's weights and can be seen as a form of regularization: These additional constraints are expected to counteract overfitting tendencies, i.e., tendencies to memorize all given mapping examples without being able to generalize. This approach has, for instance, successfully been used by Sanders and Nosofsky \cite{Sanders2018,Sanders2020}, who have fine tuned pretrained CNNs to predict the MDS coordinates on a data set of 360 rocks. In contrast to their work, we focus on the single cognitive domain of shapes and use a considerably smaller set of annotated inputs. Moreover, we consider a larger variety of machine learning setups.

\subsection{The Cognitive Domain of Shapes}
\label{Background:Shapes}

Over the past decades, there has been ample research on shape perception in different fields such as (neuro-)psychology \cite{Bar2003,OpdeBeeck2008,Biederman1987,Erdogan2017,Huang2020,Hubel1959,Li2019,Marr1978,Ons2011,Riesenhuber1999,Treisman1988}, computer vision \cite{Chella2001,Mingqiang2008,Morgenstern2021,Zhang2004}, and deep learning \cite{Baker2018,Geirhos2019,Kubilius2016,Singer2020}. Although so far no complete understanding of the shape domain has emerged, there exist some common themes that appear in multiple approaches, such as the distinction between global structure and local surface properties \cite{Baker2018,Bar2003,OpdeBeeck2008,Huang2020}, or candidate features such as \concept{aspect ratio} \cite{Bar2003,OpdeBeeck2008,Chella2001,Marr1978,Mingqiang2008,Ons2011,Treisman1988,Zhang2004}, \concept{curvature} \cite{OpdeBeeck2008,Biederman1987,Chella2001,Mingqiang2008,Ons2011,Treisman1988,Zhang2004}, or \concept{orientation} \cite{Bar2003,Chella2001,Marr1978,Hubel1959,Riesenhuber1999,Treisman1988,Zhang2004}.



In the context of conceptual spaces, G\"ardenfors \cite{Gardenfors2000} mainly refers to the model proposed by Marr and Nishihara \cite{Marr1978}, which uses configurations of cylinders to describe shapes on varying levels of granularity. This cylinder-based representation can be transformed into a coordinate system by representing each cylinder with its length, diameter, and relative location and rotation. If the number of cylinders is fixed, one can thus derive a conceptual space for the shape domain with a fixed number of dimensions. 
A related proposal for representing the shape domain within conceptual spaces has been made by Chella et al. \cite{Chella2001}, who use the more powerful class of superquadrics as elementary shape primitives, allowing them to express many simple geometric objects such as boxes, cylinders, and spheres as convex regions in their similarity space.

Both existing models of the shape domain within the conceptual spaces framework define complex shapes as a configuration of simple shape primitives and follow therefore a structural approach \cite{Erdogan2017}. The number of primitives necessary to represent a complex object may however differ between categories. Since two stimuli can therefore not necessarily be represented as two points in the same similarity space, this approach causes problems when computing distances between stimuli. Also the psychological plausibility of these approaches has so far not been established.\\

In order to provide a conceptual space representing the holistic similarity of complex shapes, Bechberger and Scheibel \cite{Bechberger2020ICCS,Bechberger2021Shapes} therefore followed a different approach: As stimuli, they used sixty line drawings of everyday objects from twelve different semantic categories (such as \concept{appliance}, \concept{bird}, \concept{building}, and \concept{insect}), taken from different sources and adjusted such that they match in relative object size as well as object position and object orientation. Six categories contained visually similar items, while the other six categories were based on visually variable items. Bechberger and Scheibel conducted a psychological study with 62 participants, where an explicit rating of the visual similarity for all pairs of items was collected. Using the averaged ratings over all participants, they then applied MDS to obtain psychological similarity spaces of different dimensionality. 

Their investigations showed that the resulting shape spaces fulfilled the predictions of the conceptual spaces framework: Distances had a high correlation to the original dissimilarities and visually coherent categories (such as \concept{appliance} and \concept{bird}) were represented as small and non-overlapping convex regions. Human ratings of the objects with respect to three psychologically motivated shape features -- namely, \concept{aspect ratio}, \concept{line curvature}, and \concept{orientation} -- could be interpreted as linear directions in these spaces. Overall, their analysis indicated that similarity spaces with three to five dimensions strike a good balance between compactness and expressiveness. We will use their four-dimensional similarity space as a target for our machine learning experiments in Section \ref{Exp}.\\

Recently, Morgenstern et al. \cite{Morgenstern2021} have proposed a 22-dimensional similarity space for shapes obtained via MDS from 109 computer vision features on a data set of 25,000 animal silhouettes. Predictions of their similarity space on novel stimuli were highly correlated with human similarity ratings ($r=0.91$), giving an indirect psychological validation to their approach. Moreover, Morgenstern et al. trained different shallow CNNs to map from original input images into their shape space. This relates their work quite strongly to our current study. However, their similarity space is based on computer vision features on silhouettes, while we start from psychological data on complex line drawings.

\section{General Methods}
\label{Methods}

In this section, we describe both our data augmentation strategy (Section \ref{Methods:Data}) and our general training and evaluation scheme (Section \ref{Methods:Training}).

\subsection{Data Augmentation}
\label{Methods:Data}


The data set of line drawings used for the psychological study by Bechberger and Scheibel \cite{Bechberger2021Shapes} is limited to 60 individual stimuli. These stimuli are all annotated with their respective coordinates in the target similarity space and are thus our main source of information for learning the mapping task. Moreover, we used 70 additional line drawings which were not part of the psychological study by Bechberger and Scheibel, but which use a similar drawing style. Most applications of convolutional neural networks focus on data sets of photographs such as ImageNet \cite{Deng2009}. In contrast to photographs, the line drawings considered in our experiments do not contain any texture or background, since they only show a single object using black lines on white ground. Sketches have similar characteristics, so we used the sketch data sets TU Berlin \cite{Eitz2012} and Sketchy \cite{Sangkloy2016} as additional data sources. From the TU Berlin corpus, we used all 20,000 sketches, while for the Sketchy corpus we selected a subset of 62,500 images by first keeping only the sketches which had been labeled as correct by the authors and then randomly selecting 500 sketches from each of the 125 categories. TU Berlin contains 250 classes and Sketchy uses 125 classes, and both data sets overlap on a subset of 98 common classes. We used the full set of 277 distinct classes when training the network on its classification objective. 


We used the following augmentation procedure to further increase the variety of inputs: For each original image, we first applied a horizontal flip with probability 0.5 and then rotated and sheared the image by an angle of up to 15 degrees, respectively. In the resulting distorted image, we identified the bounding box around the object and cropped the overall image to the size of this bounding box. The resulting cropped image was then uniformly rescaled such that its longer side had a randomly selected size between 168 and 224 pixels. Using a randomly chosen offset, the rescaled object was then put in a $224 \times 224$ image, where remaining pixels were filled with white. We used a uniform distribution over all possible resulting configurations for a given image, which makes smaller object sizes more likely since they have more translation possibilities than larger object sizes.
Please note that we did not use the augmentation steps of horizontal flips and random shears and rotations on the line drawings from the psychological study, since the similarity space contains an interpretable direction which reflects the \concept{orientation} of the object. 

For each line drawing (both from the psychological study and additional ones), we created 2000 augmented versions, while the TU Berlin data set and Sketchy were augmented with factors of 12 and 4, respectively. Overall, we obtained 120,000 data points for the line drawings from Bechberger and Scheibel, 140,000 data points for the additional line drawings, 240,000 data points for TU Berlin, and 250,000 data points for Sketchy.

\subsection{Training and Evaluation Scheme}
\label{Methods:Training}

\begin{figure}[tp]
	\centering
	\includegraphics[width = \columnwidth]{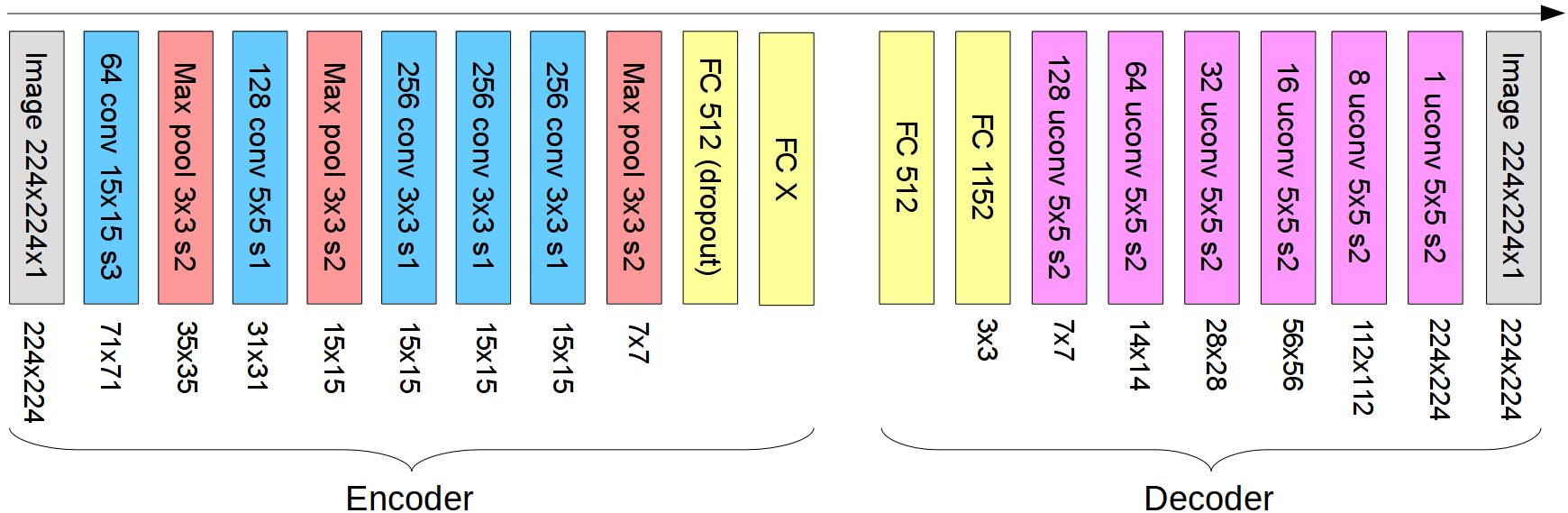}
	\caption{Structure of our CNNs (``64 conv $15 \times 15$ s3'' = convolutional layer with 64 kernels of size $15 \times 15$, using a stride of 3, ``max pool'' = max pooling layer, ``FC'' = fully connected layer, ``uconv'' = upconvolutional layer; output image size shown next to the layers).}
	\label{fig:architecture}
\end{figure}

Sketch-a-Net \cite{Yu2015,Yu2017} was the first CNN specifically designed for the task of sketch recognition and is essentially a trimmed version of AlexNet \cite{Krizhevsky2012}, the first CNN that achieved state of the art results in image classification tasks. For our encoder network (see Figure \ref{fig:architecture}), we used Sketch-a-Net and treated the size of its second fully connected layer as a hyperparameter. Moreover, we did not use dropout\footnote{Dropout is regularization technique where on each training step, a randomly chosen subset of neurons is deactivated in order to increase the network's robustness.} in this layer and used linear units instead of ReLUs\footnote{\emph{Re}ctified \emph{L}inear \emph{U}nits (ReLUs) use $max(0,x)$ as activation function and are the standard in modern CNNs, while linear units use the identity as activation function.} to allow the network to predict the MDS coordinates (which can also be negative) as part of its learned representation. Classification was realized with a softmax layer on top of the encoder (not shown in the figure). In the autoencoder setup, we additionally used a decoder network inspired by the work of Dosovitskiy and Brox \cite{Dosovitskiy2016}, which uses two fully connected layers and 6 upconvolutional layers (see Figure \ref{fig:architecture}).

We furthermore applied salt and pepper noise to the inputs before feeding them to our network. This additional noise further increases the variety of the network's inputs and can be seen as an additional form of data augmentation. We chose binary salt and pepper noise (which sets randomly selected pixels to their minimal or maximal value) rather than Gaussian noise, since the former is more adequate for our near-binary inputs where most of the pixels are either black or white.

In our experiments reported below, we trained the overall network to minimize a linear combination of the classification error (measured as softmax cross-entropy for the 277 classes), the reconstruction error (measured as sigmoid cross-entropy loss with respect to the uncorrupted images\footnote{Since our autoencoder receives a corrupted image, but needs to reconstruct the uncorrupted original, it is a so called \emph{denoising} autoencoder \cite{Vincent2008}.}) and the mapping error (measured as mean squared error for the target coordinates and the designated units of the second fully connected layer).

When evaluating the network's overall performance, we used of the following evaluation metrics: For the classification task, we report separate classification accuracies for the TU Berlin and the Sketchy data sets. For the reconstruction task, we report the reconstruction error (i.e., the binary cross-entropy loss) and for the mapping task, we report the mean squared error (MSE), the coefficient of determination $R^2$, and the mean Euclidean distance (MED) between the predicted point and the ground truth. We only use salt and pepper noise during training, but not during evaluation in order to avoid random fluctuations on the validation and test set.

Since the target coordinates used for learning and evaluating the mapping task are based only on 60 original stimuli, we decided to follow a five-fold cross validation scheme: We divided the original data points from each of the data sources into five folds of equal size and then applied the augmentation step for each fold individually. Therefore, all augmented images that were based on the same original data point are guaranteed to belong to the same fold, thus preventing potential information leaks between folds. In our overall evaluation process, we rotated through these folds, always using three folds for training, one fold for testing, and the remaining fold as a validation set for early stopping (i.e., choosing the epoch with the lowest loss). We ensured that each fold was used once for testing, once as validation set, and three times as training set. The reported numbers are always averaged across all folds. By using this five-fold cross-validation technique, we implicitly trained five neural networks with the same hyperparameter settings, but slightly different data. Our averaged results therefore approximate the expected value of the neural network's performance on unseen inputs and hence the generalizability of the learned mapping.

During training, we use the Adam optimizer \cite{Kingma2014} as a variant of stochastic gradient descent, with the initial learning rate set to $0.0001$, the default parameter settings of $\beta_1 = 0.9$, $\beta_2 = 0.999$, $\epsilon = 10^{-8}$, and a batch size\footnote{The batch size determines how many training examples are used for computing a single parameter update.} of 128. We ensure that each minibatch contains examples from all relevant data sources according to their relative proportions: When training only on the classification task, we take 63 examples from TU Berlin and 65 from Sketchy. When training on both the classification and the mapping task, we use 25 line drawings, 51 sketches from TU Berlin, and 52 examples from Sketchy. Whenever the reconstruction task is involved, we use 21 line drawings, 24 additional line drawings, 41 examples from TU Berlin, and 42 data points from Sketchy. We always train the network for 200 full epochs\footnote{One epoch is one full iteration over the complete training set.} and select the epoch with the lowest validation set loss (classification loss or reconstruction loss for the pretraining experiments, and mapping loss for the multi-task learning experiments) in order to compute performance on the test set.

\section{Experiments}
\label{Exp}

In this section, we report the results of the experiments carried out with our general setup as described in Section \ref{Methods}. In Section \ref{Exp:Pretraining}, we train our network exclusively on the classification and reconstruction task, respectively. This provides a starting point for our transfer learning experiments in Section \ref{Exp:TransferLearning}, where we apply a linear regression on top of the pretrained CNNs. In Section \ref{Exp:MultiTaskLearning}, we then follow a multi-task learning approach, where both the mapping task and the secondary objective are optimized jointly. Finally, in Section \ref{Exp:Generalization}, we investigate how well the different approaches generalize to target similarity spaces of varying dimensionality.

\subsection{Pretraining}
\label{Exp:Pretraining}

We first considered a default setup of the hyperparameters based directly on Sketch-a-Net \cite{Yu2015,Yu2017} and AlexNet \cite{Krizhevsky2012}: We used a weight decay\footnote{A weight decay term penalizes the magnitude of the network's weights.} of 0.0005, dropout in the first fully connected layer, and a representation size of 512 neurons in the second fully connected layer. Moreover, we used 10\% salt and pepper noise during training. For the decoder network, we use neither dropout nor weight decay.
As evaluation metrics for the classification task, we considered the accuracies reached on TU Berlin and Sketchy, while for the autoencoder, the reconstruction error was used. In both cases, we also computed the monotone correlation of distances in the feature space to the dissimilarity ratings of Bechberger and Scheibel \cite{Bechberger2021Shapes}, measured with Kendall's $\tau$ \cite{Kendall1938}. Since a full grid search on many candidate values per hyperparameter was computationally prohibitive (especially in the context of a cross validation), we first identified up to two promising settings for each hyperparameter for both network types, before conducting a small grid search on the remaining combinations. The configurations selected in this grid search are shown in Table \ref{tab:pretraining}.

\begin{table}[t]
 	\caption{Selected hyperparameter configurations.}
 	\centering
 	\setlength\tabcolsep{0.15cm}
 	\begin{tabular}{|c||c|c|c|c||c|c|}
    	\hline
		\multirow{3}{*}{\textbf{Configuration}} & \multicolumn{4}{c||}{\textbf{Encoder}} &  \multicolumn{2}{c|}{\textbf{Decoder}} \\ 
		& \textbf{Weight} & \multirow{2}{*}{\textbf{Dropout}} & \textbf{Noise} & \textbf{Rep.} & \textbf{Weight} & \multirow{2}{*}{\textbf{Dropout}} \\ 
		& \textbf{Decay} & & \textbf{Level} & \textbf{Size} & \textbf{Decay} & \\ \hline \hline
		
		$C_\concept{default}$ 		& 0.0005	& True 	& 10\%	& 512	& --	& --	\\ \hline
		$C_\concept{small}$ 		& 0.0005	& True 	& 10\%	& 256	& --	& --	\\ \hline
		$C_\concept{correlation}$	& 0.001		& False & 10\%	& 512	& --	& --	\\ \hline \hline
		
		$R_\concept{default}$		& 0.0005	& True 	& 10\%	& 512	& 0.0	& False	\\ \hline
		$R_\concept{best}$			& 0.0		& False	& 10\%	& 512	& 0.0	& False	\\ \hline 

  	\end{tabular}
	\label{tab:pretraining}
\end{table}

For the classifier network, the best classification performance (with accuracies of 63.2\% and 79.3\% on TU Berlin and Sketchy, respectively) was obtained by our default setup $C_\concept{default}$. This is considerably lower than the 77.9\% on TU Berlin reported for the original Sketch-a-Net \cite{Yu2017}, which however used a much more sophisticated data augmentation and pretraining scheme.
A considerably higher correlation of $\tau \approx 0.33$ (instead of $\tau \approx 0.27$ for $C_\concept{default}$) to the dissimilarity ratings could be obtained by disabling dropout and increasing the weight decay ($C_\concept{correlation}$), however at the cost of considerably reduced classification accuracies of 36.4\% and 61.5\% on TU Berlin and Sketchy, respectively. Since reducing the representation size barely affected classification performance, we also consider $C_\concept{small}$, which uses 256 units and otherwise default parameters.

For the autoencoder, we observed that completely disabling both weight decay and dropout in both the encoder and the decoder let to considerably improved reconstruction performance (reconstruction error of 0.08 for $R_\concept{best}$ in comparison to 0.13 for $R_\concept{default}$). Also the correlation to the dissimilarities increased from $\tau \approx 0.22$ to $\tau \approx 0.30$. Manipulation of all other hyperparameters did not lead to further improvements.

\subsection{Transfer Learning}
\label{Exp:TransferLearning}

For our transfer learning task, we extracted the hidden representation of each network configuration for each of the augmented line drawings. We trained a linear regression from these feature spaces to the four-dimensional shape space by Bechberger and Scheibel \cite{Bechberger2021Shapes}. In addition to the linear regression, we also consider a lasso regression (which introduces an additional L1 penalty on the model's weights) with the following settings for the regularization strength $\beta$:
$$\beta \in \{0.001, 0.002, 0.005, 0.01, 0.02, 0.05, 0.1, 0.2, 0.5, 1.0, 2.0, 5.0, 10.0\}$$

\begin{table}[t]
 	\centering
 	\setlength\tabcolsep{0.05cm}
	\caption{Results of our experiments on the four-dimensional target space. The respective best values for each configuration are shown in boldface.}
 	\begin{tabular}{|c|c|c|c||c|c|c|c|}
    	\hline
		\textbf{Configuration} & \textbf{Task} & \textbf{Regressor} & \textbf{$\beta/\lambda$} & \textbf{$\tau$} & \textbf{MSE} & \textbf{MED} & \textbf{$R^2$} \\ \hline \hline
    	
    	Any			& Any			& Zero Baseline		& --
    						 	& --	& 1.0000 	& 0.9940 	& 0.0000 \\ \hline \hline

    	\multirow{3}{*}{$C_\concept{default}$}
    				& \multirow{2}{*}{Transfer}		
    								& Linear	& --
    							& 0.2743	& 0.5567 	& 0.6879	& 0.4409 \\ \cline{3-8}
    				& 				& Lasso 	& 0.05
    							& 0.2743	& 0.4775 	& 0.6419 	& 0.5216 \\ \cline{2-8}
    				& Multi-Task	& CNN		& 0.0625
    							& \textbf{0.4141}	& \textbf{0.4041}	& \textbf{0.5920}	& \textbf{0.5775}			\\ \hline  \hline

    	\multirow{3}{*}{$C_\concept{small}$}
    				& \multirow{2}{*}{Transfer}		
    								& Linear		& --
    							& 0.2777	& 0.5373 	& 0.6737 	& 0.4575 \\ \cline{3-8}
    				& 				& Lasso 	& 0.02
    							& 0.2777	& 0.4737 	& 0.6396	& 0.5246 \\ \cline{2-8} 
    				& Multi-Task	& CNN		& 0.125		
    							& \textbf{0.4118}	& \textbf{0.4182}	& \textbf{0.6020}	& \textbf{0.5567} 	\\ \hline \hline

    	\multirow{3}{*}{$C_\concept{correlation}$}
    				& \multirow{2}{*}{Transfer}		
    								& Linear		& --
    							& 0.3292	& 0.7307 	& 0.7825 	& 0.2624 \\ \cline{3-8}
    				& 				& Lasso 	& 0.05
    							& 0.3292	& 0.5478 	& 0.6815 	& 0.4505 \\ \cline{2-8}
    				& Multi-Task	& CNN		& 2.0		
    							& \textbf{0.4534}	& \textbf{0.4513}	& \textbf{0.6115}	& \textbf{0.5201}				\\ \hline \hline
    	
    	\multirow{3}{*}{$R_\concept{default}$}
    				& \multirow{2}{*}{Transfer}		
    								& Linear		& --
    							& 0.2228 	& 0.9709 	& 0.9054 	& 0.0168 \\ \cline{3-8}
    				& 				& Lasso 	& 0.02, 0.05
    							& 0.2228 	& 0.8315 	& 0.8739 	& 0.1631 \\ \cline{2-8} 
    				& Multi-Task	& CNN		& 2.0		
    							& \textbf{0.3533} 	& \textbf{0.6211}	& \textbf{0.7297}	& \textbf{0.3369}			\\ \hline \hline

    	\multirow{5}{*}{$R_\concept{best}$}
    				& \multirow{2}{*}{Transfer}		
    								& Linear		& --
    							& 0.3019	& 1.0791 	& 0.9362 	& -0.0886 \\ \cline{3-8}
    				& 				& Lasso 	& 0.02
								& 0.3019	& 0.7376	& 0.8102	& 0.2605 \\ \cline{2-8}
    				& \multirow{3}{*}{Multi-Task}		
    								& \multirow{3}{*}{CNN}		
    										& 0.25, 
    							& \multirow{2}{*}{\textbf{0.4033}} 	& \multirow{2}{*}{\textbf{0.5494}}	& \multirow{2}{*}{\textbf{0.6846}}	& \multirow{2}{*}{\textbf{0.4213}} \\
    				&				&		& 0.5, 2.0  & & & & \\ \cline{4-8}
    				& 				& 			& 0.0625
    							& 0.3893 	& 0.5504	& 0.6851	& 0.4144 	\\ \hline 

  	\end{tabular}
	\label{tab:results}
\end{table}

Table \ref{tab:results} contains the results of these regression experiments. As we can see, the linear regression performs considerably better than the zero baseline (which always predicts the origin of the target space) for the classification-based feature spaces, but not for the reconstruction-based feature spaces. Moreover, regularization helps to improve performance on all feature spaces. A lasso regression on $C_\concept{small}$ slightly outperforms $C_\concept{default}$, hinting at an advantage of smaller representation sizes. $C_\concept{correlation}$ does not yield competitive results, indicating that classification accuracy is a more useful selection criterion in pretraining than the correlation to human dissimilarity ratings.

Overall, transfer learning based on classification networks seems to be much more successful than transfer learning based on autoencoders, even when considering a lasso regressor. The reason for the relatively poor performance of $R_\concept{best}$ and $R_\concept{default}$ can be seen in Table \ref{tab:transfer_clusters}, where we analyze how well the different augmented versions of the shape stimuli from Bechberger and Scheibel \cite{Bechberger2021Shapes} are separated in the different feature spaces. We used the Silhouette coefficient \cite{Rousseeuw1987}, where larger values indicate a clearer separation of clusters. As we can see, the different augmented versions of the same original line drawing do not form any notable clusters in the reconstruction-based feature space. On the other hand, a relatively strong clustering can be observed for classification-based feature spaces under both noise conditions, indicating that the network is able to successfully filter out noise. We assume that this difference is based on the fact that the autoencoder needs to preserve very detailed information about its input (both local and global shape information) in order to create a faithful reconstruction, while a classification network only needs to preserve pieces of information that are highly indicative of class membership (rather global than local information).

\begin{table}[t]
 	\caption{Cluster analysis of the augmented images in the individual feature spaces (averaged across all folds) using the Silhouette coefficient and the Cosine distance.}
 	\centering
 	\setlength\tabcolsep{0.1cm}
 	\begin{tabular}{|c||c|c|c||c|c|}
    	\hline
		\textbf{Configuration} 		& $C_\concept{default}$ 	& $C_\concept{small}$ 		& $C_\concept{correlation}$ 	& $R_\concept{default}$ & $R_\concept{best}$	\\ \hline \hline
		\textbf{0\% Noise}			& 0.6448	& 0.6347	& 0.5310	& -0.0359 & 0.0818\\ \hline
		\textbf{10\% Noise}			& 0.6364	& 0.6263	& 0.5180 	& -0.0300 & 0.0768\\ \hline
  	\end{tabular}
	\label{tab:transfer_clusters}
\end{table}

\subsection{Multi-Task Learning}
\label{Exp:MultiTaskLearning}

In our multi-task learning experiments, we trained our networks in the different configurations again from scratch, using however also the mapping loss as additional training objective. Instead of a two-phase process as used in the transfer learning setup, we therefore optimize both objectives at once. 
When training the networks, we varied the relative weight $\lambda$ of the mapping loss in order to explore different trade-offs between the two tasks. We explored the following settings (where $\lambda = 0.25$ approximately reflects the relative proportion of mapping examples in the classification task):
$$\lambda \in \{0.0625, 0.125, 0.25, 0.5, 1.0, 2.0\}$$

Table \ref{tab:results} also contains the results of our multi-task learning experiments. As we can observe, mapping performance is considerably better in the multi-task setting than in the transfer learning setting for all of the configurations under investigation. The best results are obtained for $C_\concept{default}$, which is followed closely by $C_\concept{small}$. $C_\concept{correlation}$ performs again considerably worse than the other classification-based setups, although its best multi-task results are still superior to all transfer learning results. Moreover, both reconstruction-based setups are not able to close the performance gap to the classification-based networks also under multi-task learning. These observations indicate that the multi-task learning regime is more promising than the transfer learning approach and that classification is a more helpful secondary task than reconstruction.

When taking a closer look at the optimal values for $\lambda$, we note that for both the $C_\concept{default}$ and the $C_\concept{small}$ setting, relatively small values of $\lambda \in \{0.0625, 0.125\}$ have been selected. 
For the $C_\concept{correlation}$ configuration, however, a relatively large mapping weight of $\lambda = 2.0$ leads to the best mapping results, indicating that this configuration requires stronger regularization than others.
Also for $R_\concept{default}$, a relatively large mapping weight of $\lambda = 2.0$ yielded the best performance, while no unique best setting for $\lambda$ could be determined for the $R_\concept{best}$ configuration, where different metrics are optimized by different hyperparameter settings -- here, $\lambda = 0.0625$ provides a reasonable trade-off.

In all cases, the introduction of the mapping loss leads to a considerable increase in the correlation $\tau$ to the dissimilarity ratings. This effect is however to be expected, since the mapping loss tries to align a part of the internal representation with the coordinates of the similarity space, which is explicitly based on the psychological dissimilarity ratings.

\subsection{Generalization to Other Target Spaces}
\label{Exp:Generalization}


So far, we have only considered a four-dimensional target space. In this section, we investigate how well the different approaches generalize to target spaces of different dimensionality. We considered the respective best setups for all combinations of classification-based vs. reconstruction-based networks and transfer learning vs. multi-task learning (cf. Table \ref{tab:results}) and retrained them (using the same values of $\beta/\lambda$) on all other target spaces (one to ten dimensions) of Bechberger and Scheibel \cite{Bechberger2021Shapes}, using again a five-fold cross validation.\\

\begin{figure}[tp]
	\centering
	\includegraphics[width=\columnwidth]{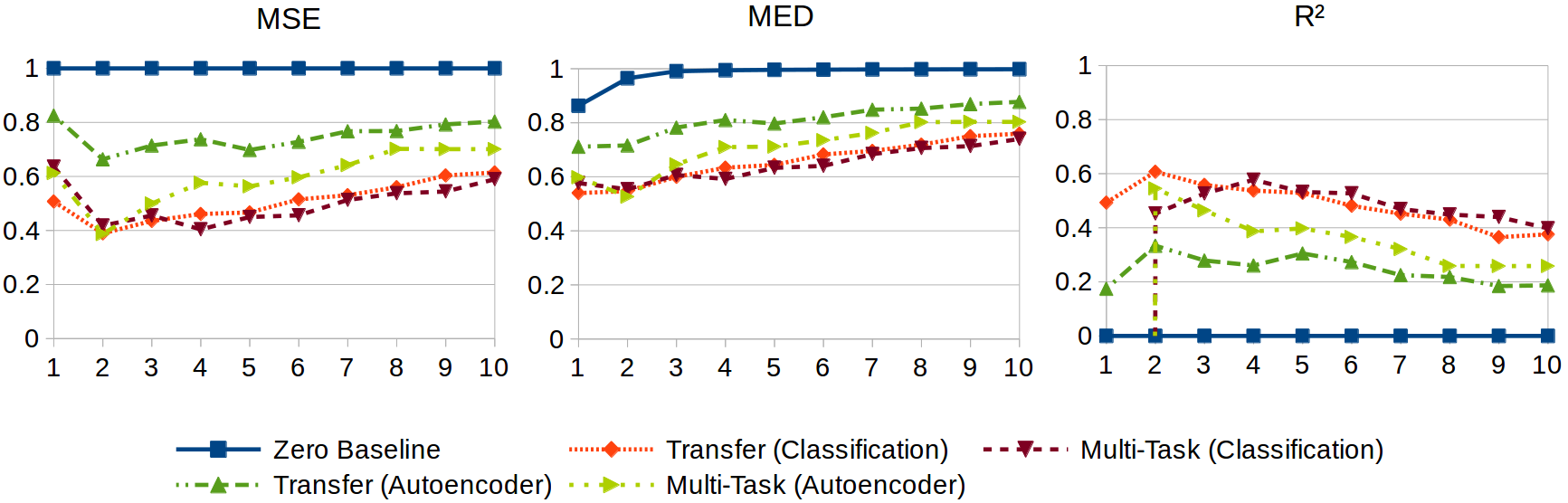}
	\caption{Results of our generalization experiments.}
	\label{fig:generalization_results}
\end{figure}

Figure \ref{fig:generalization_results} illustrates the results of these generalization experiments for our three evaluation metrics. Both transfer learning approaches reach their peak performance for a two-dimensional target space, even though they have been optimized on the four-dimensional similarity space. Only with respect to the MED, performance is best on the one-dimensional target space. However, also the MED of the zero baseline is smallest for a one-dimensional space. If we consider the relative MED (by dividing through the MED of the zero baseline), then the best performance is again obtained on a two-dimensional target space. In all cases, classification-based transfer learning is clearly superior to reconstruction-based transfer learning.

The multi-task learners on the other hand do not show such a uniform pattern: While the reconstruction-based approach also obtains its optimum for a two-dimensional target space, the classification-based multi-task learner seems to prefer a four-dimensional target space. Moreover, both multi-task learners are more sensitive to the dimensionality of the target space than the transfer learning approaches: The classification-based multi-task learner considerably outperforms all other approaches on medium- to high-dimensional target spaces, while falling behind for a smaller number of dimensions. The reconstruction-based multi-task learner on the other hand performs quite poorly on high-dimensional spaces while becoming competitive on low-dimensional target spaces. Both multi-task learners use a mapping weight of $\lambda = 0.0625$, i.e., the smallest value we investigated. However, the size of the classification and reconstruction loss has differed considerably, with a classification loss of around 1.3 to 1.6, compared to a reconstruction loss of 0.10 to 0.12 (both measured on the test set). The relative influence of the mapping objective on the overall optimization is thus considerably greater in the classification-based multi-task learner. One may therefore speculate that even smaller values of $\lambda$ would have benefited the classification-based multi-task learner for smaller target spaces.\\

Overall, the results of this generalization experiment confirm the effects reported in our earlier study \cite{Bechberger2021NOUN}, where we also observed a performance sweet spot for a two-dimensional target space in a transfer learning setting. Again, we can argue that this strikes a balance between a clear semantic structure in the target space and a small number of output variables to predict. The observed sensitivity of the multi-task learning approach indicates that the target space should be carefully chosen before optimizing the multi-task learner. 

\section{Discussion and Conclusion}
\label{Conclusion}

In this paper, we have aimed to learn a mapping from line drawings to their corresponding coordinates in a psychological shape space. We have compared classification-based networks to autoencoders, investigating both transfer learning and multi-task learning. Overall, classification seemed to be a better secondary task than reconstruction, and multi-task learning consistently outperformed transfer learning. We found that the best performance in general was reached for classification-based multi-task learning, but that this approach was quite sensitive to the dimensionality of the target space.

We can compare our results to our earlier study \cite{Bechberger2021NOUN}, where we used a lasso regression on top of a CNN pretrained to classify photographs. There, we achieved for a four-dimensional target space a MSE of about 0.59, a MED of about 0.73, and a coefficient of determination of $R^2 \approx 0.39$. These numbers are considerably worse than the ones obtained for classification-based transfer learning Section \ref{Exp:TransferLearning}, indicating that the shape space considered in the current study poses an easier regression problem. Moreover, we can compare our performance with respect to the coefficient of determination to the results reported by Sanders and Nosofsky \cite{Sanders2018}, who reported a value of $R^2 \approx 0.77$ for an eight-dimensional target space and a more complex network architecture, using a data set of 360 stimuli. Our best results with $R^2 \approx 0.61$ on a two-dimensional target space are considerably worse than this and clearly not good enough for practical applications. We assume that performance in our scenario is heavily constrained by the network size and the number of stimuli for which dissimilarity ratings were collected. This urges for further experimentation with more complex architectures, larger data sets, different augmentation techniques, and additional regularization approaches.\\


Overall, our present study has illustrated that it is in principle possible to predict the coordinates of a given input image in a psychological similarity space for the shape domain. Although performance is not yet satisfactory, this is an important step towards making conceptual spaces usable for cognitive AI systems.
Once a robust mapping of reasonably high quality has been obtained, one can use the full expressive power of the conceptual spaces framework: For instance, the interpretable directions reported by Bechberger and Scheibel \cite{Bechberger2021Shapes} can give rise to an intuitive description of novel stimuli based on psychological features. Also categorization based on conceptual regions, common-sense reasoning strategies, and concept combination can then be implemented on top of the predicted coordinates in shape space.

The approach presented in this article can of course also be generalized to other domains and data sets such as the THINGS data base and its associated embeddings \cite{Hebart2020} or the recently published similarity ratings and embeddings for a subset of ImageNet \cite{Roads2020}. It can furthermore be seen as a contribution to the currently emerging field of research which tries to align neural networks with psychological models of cognition \cite{Attarian2020, Battleday2020,Battleday2021,Jha2020,Kubilius2016,Lake2015,Morgenstern2021,Peterson2017,Peterson2018,Peterson2019,Sanders2018,Sanders2020,Singh2020,Sorscher2021}.
%
%
%
\bibliographystyle{splncs04}
\bibliography{ref.bib}

\end{document}